# Multi-task Sentence Encoding Model for Semantic Retrieval in Question Answering Systems

Qiang Huang
Big Data Department
Baidu Inc.
Shenzhen, China
huangqiang03@baidu.com

Jianhui Bu
Big Data Department
Baidu Inc.
Beijing, China
bujianhui@baidu.com

Weijian Xie
Big Data Department
Baidu Inc.
Shenzhen, China
xieweijian@baidu.com

Shengwen Yang
Big Data Department
Baidu Inc.
Beijing, China
yangshengwen@baidu.com

Weijia Wu
Big Data Department
Baidu Inc.
Shenzhen, China
wuweijia01@baidu.com

Liping Liu
Big Data Department
Baidu Inc.
Beijing, China
liuliping@baidu.com

*Abstract*—Question Answering (QA) systems are used to provide proper responses to users' questions automatically. Sentence matching is an essential task in the QA systems and is usually reformulated as a Paraphrase Identification (PI) problem. Given a question, the aim of the task is to find the most similar question from a QA knowledge base. In this paper, we propose a Multi-task Sentence Encoding Model (MSEM) for the PI problem, wherein a connected graph is employed to depict the relation between sentences, and a multi-task learning model is applied to address both the sentence matching and sentence intent classification problem. In addition, we implement a general semantic retrieval framework that combines our proposed model and the Approximate Nearest Neighbor (ANN) technology, which enables us to find the most similar question from all available candidates very quickly during online serving. The experiments show the superiority of our proposed method as compared with the existing sentence matching models.

*Keywords—Question Answering systems, sentence matching, encoding model, multi-task learning, semantic retrieval framework*

## I. INTRODUCTION

Question answering systems have been widely studied in both the academic and industrial community and are widely applied to various scenarios. There are full-blown applications like Amazon's Alexa, Apple's Siri, Baidu's DuerOS, Google's Assistant and Microsoft's Cortana. Generally, there are two types of question answering systems: (1) information retrieval-based (IR-based) [1], and (2) generation-based [2]. In this work, we focus on building an IR-based QA system to answer the Frequently Asked Questions (FAQ). The critical part of IR-based QA system is to find the most similar question from a massive QA knowledge base, which could be further reformulated as a Paraphrase Identification (PI) problem, also known as sentence matching.

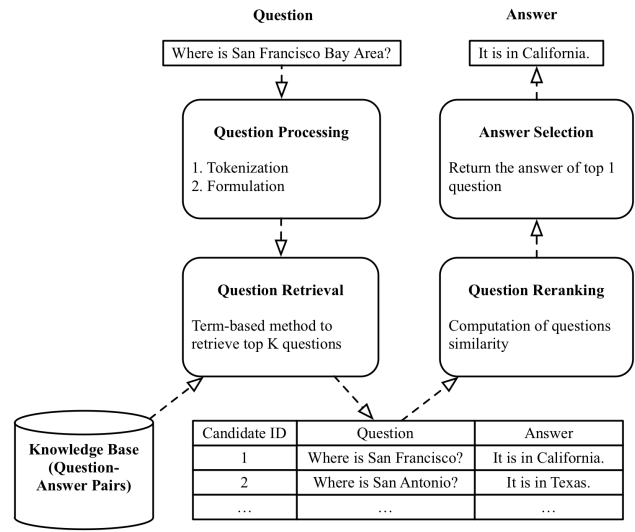

Fig. 1. The workflow of the traditional IR-based QA systems.

In recent years, neural network models have achieved great success in sentence matching. Depends on whether to use cross-sentence features or not, sentence-matching models can be classified roughly into two types: (1) encoding-based, and (2) interaction-based. It is generally accepted that the interaction-based models could get better performance than the encoding-based models on certain datasets because they have abundant interaction features. However, the leaderboards of published large datasets such as SNLI [3] and MultiNLI [4] encourage conducting research on the encoding-based models around the semantic representation, because the encoding-based models can learn vector representations of individual sentences, which can be further applied to other natural language processing tasks.

Models in practical QA systems have two main disadvantages. Firstly, they consider the semantic sentence matching as a binary classification problem, assuming that samples are independent of one another as default. However, the paraphrase relation between sentences could be transmitted. For example, if $question_1$ and $question_2$ are paraphrases, and $question_2$ and $question_3$ are paraphrases, we can infer that $question_1$ and $question_3$ are also paraphrases. Secondly, because of the hard time delay constraint in the online prediction procedure of a traditional IR-based QA system, as shown in Fig. 1, existing models often play the role of a re-rank module that depends on the results from the question analysis and recall module. Thus they could only re-rank a few candidates from term-based index recall modules like Lucene, instead of retrieving the most similar question from all candidates.

In this paper, we aim to address these two challenges. The main contributions of this work are summarized as follows:

- We employ a connected graph to depict the paraphrase relation between sentences for the PI task, and propose a multi-task sentence-encoding model, which solves the paraphrase identification task and the sentence intent classification task simultaneously.
- We propose a semantic retrieval framework that integrates the encoding-based sentence matching model with the approximate nearest neighbor search technology, which allows us to find the most similar question very quickly from all available questions, instead of within only a few candidates, in the QA knowledge base.

We evaluated our proposed method on various QA datasets and the experimental results show the effectiveness and superiority of our method. First, it proves that we can achieve better performance with multi-task learning. Besides, our method can achieve state-of-the-art performance compared with existing encoding-based models and interaction-based models.

## II. RELATED WORKS

### A. Natural Language Sentence Matching

Natural language sentence matching (NLSM) has gone through substantial developments in recent years. It plays a central role in a large number of natural language processing tasks. For the paraphrase identification (PI) task, NLSM is utilized to determine whether two sentences are paraphrases or not.

The developments of deep neural networks and the emergence of large-scale annotated datasets have led great progress on NLSM tasks. Depending on whether the cross-sentence features or attention from one sentence to another were used, two types of deep neural network models have been proposed for NLSM. The first type of models is encoding-based, where sentences are encoded into sentence vectors without any cross interaction, then the matching decision is made solely based on the two sentence vectors. Typical representatives of such methods include Stack-augmented Parser-Interpreter Neural Network (SPINN) [5], Shortcut-Stacked Sentence

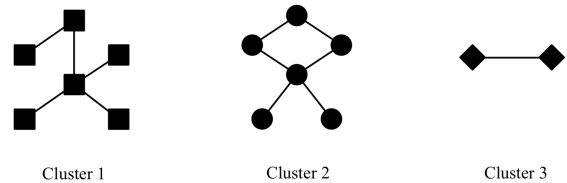

Fig. 2. The sentence clusters with similar intent.

Encoders (SSE) [6], or Gumbel Tree-LSTM [7]. The other type of methods, called interaction-based model, make use of cross interaction of small units (such as words) to express word-level or phrase-level alignments for performance improvements. The main representatives are Enhanced Sequential Inference Model (ESIM) [8], Bilateral Multi-Perspective Matching Model (BiMPM) [9], and Densely Interactive Inference Network (DIIN) model [10]. Generally, the second type of methods captures more interactive features between the two input sentences, so it can achieve better performance. On the other hand, the encoding-based model is much smaller and easier to train, and the vector representations can be further used for sentence clustering, semantic search, visualization and many other tasks. The advantages of encoding-based models are much more significant to QA systems in the industry, so we focus on the research of encoding-based models.

### B. Approximate Nearest Neighbor

Approximate nearest neighbor (ANN) search has been a hot topic over decades and plays an important role in machine learning, computer vision and information retrieval etc. For dense real-valued vectors, such as vector representations of images or natural languages, many data structures and algorithms have been proposed to improve the retrieval efficiency of ANN search. There are four types of mainstream methods, including tree structure based [11], hashing based [12], quantization based [13], and graph based [14]. In the task of image retrieval, ANN index technologies have been used to build efficient image retrieval systems.

In this paper, inspired by research on image retrieval systems, we propose a semantic retrieval framework for QA systems, which combines ANN search technology and sentence encoding technology.

## III. PROPOSED METHOD

### A. The Overall Architecture

Motivated by the fact that the questions in a large QA knowledge base are not independent, we try to utilize the paraphrase relationship among questions to facilitate modeling of question representation. Each sentence is regarded as a vertex and an edge is added between a pair of vertices if they are paraphrases. In this way we can build an undirected graph to represent the paraphrase relationship among sentences, wherein a connected sub-graph can be seen as a sentence cluster with similar intent, as shown in Fig. 2. On this basis, we could train a multi-class classification model for sentence intent classification. Two sentences can be considered as paraphrases if they are classified into the same class by the model.

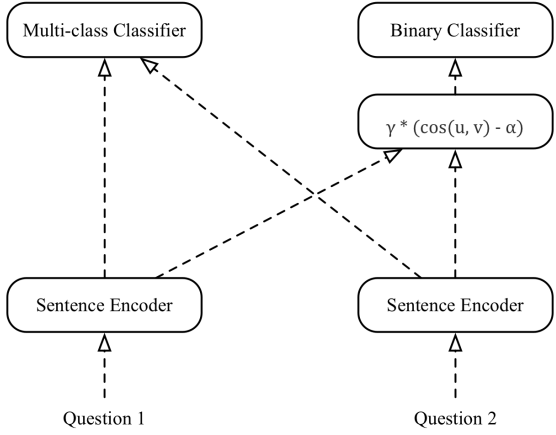

Fig. 3. Overall architecture of the proposed model.

As Fig. 3 shows, we employ a multi-task learning method to simultaneously train a sentence matching model and a sentence intent classification model by sharing the sentence encoder between two tasks. Take Quora Question Pairs dataset [15] as an instance: input data question₁ and question₂ will be encoded as sentence representation $u, v$ by the sentence encoder.

On the one hand, we use a softmax layer and cross entropy loss function to train the multi-class sentence intent classification model as follows:

$$L_u = -\sum_{C=1}^{M} y_C * \log(\text{softmax}(W^T u + b)) \quad (1)$$

$$L_v = -\sum_{C=1}^{M} y_C * \log(\text{softmax}(W^T v + b)) \quad (2)$$

$C$ in (1) & (2) is the class index of sentence intent, and M represents the number of classes.

On the other hand, for the convenience of integrating with approximate nearest neighbor (ANN) libraries, which only support cosine distance, Euclidean distance, Manhattan distance, Hamming distance, or Dot (Inner) Product distance, we use a simple cosine matching layer instead of a more complicated multi-layer perceptron as in [16].

$$\hat{y} = \text{sigmoid}(\gamma * (cos(u, v) - \alpha)) \quad (3)$$

$$L_{match} = -(y * \log \hat{y} + (1 - y) * \log(1 - \hat{y})) \quad (4)$$

In (3), $\gamma$ and $\alpha$ are hyperparameters. Equation (4) is loss function of the matching layer. The overall loss function is as follows:

$$loss = \lambda * L_{match} + \frac{(1-\lambda)*(L_u+L_v)}{2} \quad (5)$$

where $\lambda$ in (5) is a hyperparameter for balancing the loss of each task in the multi-task learning and $0 \leqslant \lambda \leqslant 1$.

*B. Sentence Encoder*

The overall architecture of our sentence encoder is illustrated in Fig. 4. The encoder transforms the input sentence into a fixed-length embedding. The details of each component in the sentence encoder will be described in the subsections that follow.

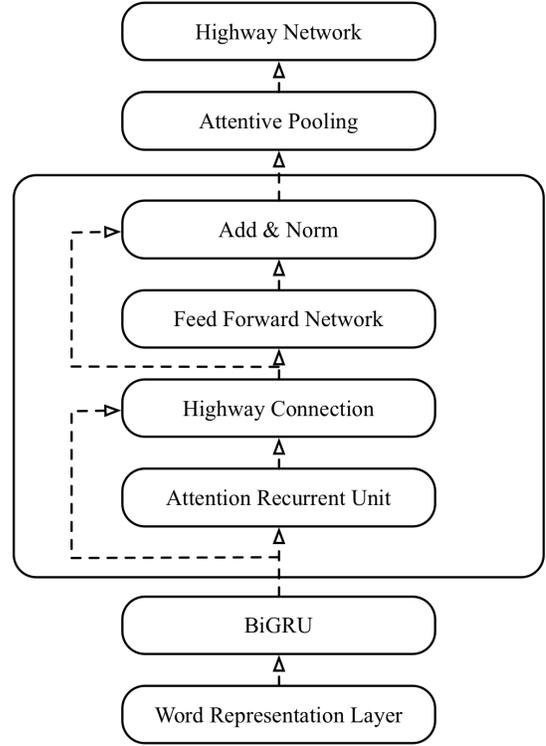

Fig. 4. Sentence Encoder.

*C. Word Representation Layer*

Word Representation Layer is consists of word embedding and character representation. For word embedding, we use pre-trained GloVe word embeddings [17] to represent each word as a *d*-dimensional vector. For character representation, we infuse randomly initialized values to max-pooling convolution layer to compute the character representation of each word. We concatenate the word embedding and character representation to get the final word representation.

*D. Bidirectional Gated Recurrent Unit*

We use Bidirectional Gated Recurrent Unit (BiGRU) [18] to maintain the sequential information about the sentence being modeled. BiGRU is consists of a forward directional GRU and a backward directional GRU. Forward directional GRU process inputs sequence from left to right, backward directional GRU on the contrary. Let us describe how the $t_{th}$ hidden unit is computed:

$$r_t = \text{sigmoid}(x_t W_r^1 + h_{t-1} W_r^2) \quad (6)$$

$$f_t = \text{sigmoid}(x_t W_f^1 + h_{t-1} W_f^2) \quad (7)$$

$$\tilde{h}_t = \tanh(x_t W_h^1 + (h_{t-1} \circ r_t) W_h^2) \quad (8)$$

$$h_t = (1 - f_t) \circ \tilde{h}_t + f_t \circ h_{t-1} \quad (9)$$

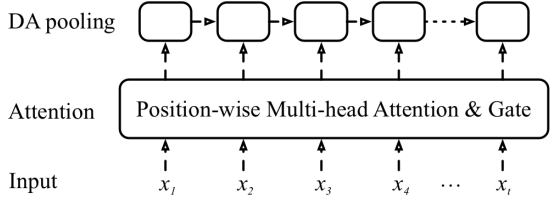

Fig. 5. Attention Recurrent Unit (ARU).

∘ in (8) & (9) denotes the element-wise product. In GRU cell, context information of $t_{th}$ hidden unit is carried over by the last hidden unit state $h_{t-1}$.

*E. Attention Recurrent Unit*

We propose an Attention Recurrent Unit (ARU) based on the attention mechanism. As shown in Fig. 5, the term "Gate" stands for the element-wise forget gate $f$. We apply position-wise multi-head attention to compute the context representation $C$ (self-attention computed here), and use it to compute the linear transformation $\tilde{h}$ with the input $x$ and the forget gate $f$:

$$C = \text{PosMultiHead}(x, x, x) \quad (10)$$

$$f = \text{sigmoid}(xW_f^1 + CW_f^2) \quad (11)$$

$$\tilde{h} = \tanh(xW_h^1 + CW_h^2) \quad (12)$$

Position-wise multi-head attention is a variant of multi-head attention [19] as shown below:

$$\text{PosAttention}(Q, K, V) = \text{softmax}\left(\frac{QK^T}{\sqrt{d_k}} + M_{pos}\right)V \quad (13)$$

$$\text{PosMultiHead}(Q, K, V) = \text{concat}(head_1, \ldots, head_n) \quad (14)$$

where $head_i = \text{PosAttention}(QW_i^Q, KW_i^K, VW_i^V)$.

$M_{pos} \in \mathbb{R}^{n \times u}$ is a weight matrix where n is the number of words on sentence, which will be updated during training.

We use a dynamic average (DA) pooling [20] to improve the sequence encoding capability further:

$$h = (1 - f_t) \circ \tilde{h}_t + f_t \circ h_{t-1} \quad (15)$$

After that, we use a highway connection [21] to connect input and output:

$$r = \text{sigmoid}(xW_r^1 + CW_r^2) \quad (16)$$

$$o = (1 - r) \circ x + r \circ h \quad (17)$$

Compared with GRU, the computation of the $t_{th}$ hidden unit is no more depending on the last hidden unit state $h_{t-1}$, so that (10), (11), (12) could be computed in parallel on GPU effectively and (15) just needs a simple computation. In this section, we also refer to the works of Quasi-RNN [22] and SRU [23].

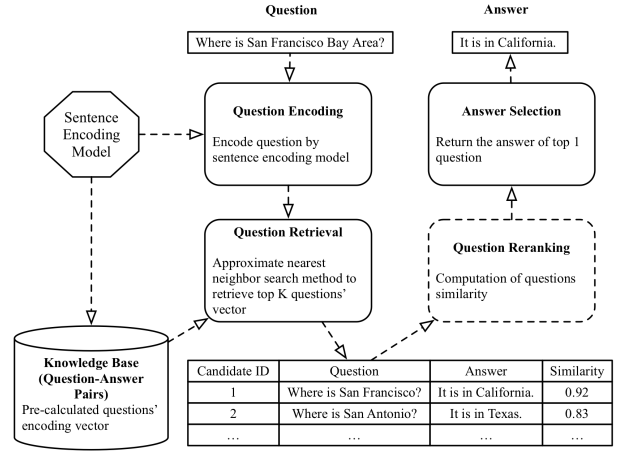

Fig. 6. The workflow of the semantic retrieval QA systems.

*F. Feed-Forward Network*

We use a feed-forward network same as the Transformer [19]. It uses a multi-layer perceptron with two layers and uses activation function ReLU, as follows:

$$\text{FFN}(o) = max(0, oW_1 + b_1)W_2 + b_2 \quad (18)$$

FFN function is applied to each output state of ARU. The term "Add" in Fig. 4 represents the residual connection [24] and the term "Norm" represents the layer normalization [25]. The output of FFN would be incorporated with the Add & Norm layer to simplify the network's optimization.

*G. Attentive Pooling*

We perform an attentive pooling operation [26] over the output of the FFN, which would convert them into a fixed-length vector. It can be formulated as follows:

$$A = \text{softmax}(W_{s2}\tanh(W_{s1}o^T)) \quad (19)$$

$$M = Ao^T \quad (20)$$

In (20), $o \in \mathbb{R}^{n \times u}$ is the output of FFN, where $n$ is the number of words in the sentence and $u$ is the number of hidden units of ARU. $W_{s1} \in \mathbb{R}^{u \times d_a}$ and $W_{s2} \in \mathbb{R}^{d_a \times r}$ are two weight matrices where $d_a$ and $r$ are hyperparameters that could be set manually. After the attentive pooling layer, the output matrix $M \in \mathbb{R}^{r \times u}$ consists of $r$ sentence representations with $u$-dimensional vectors. We concatenate the above $r$ vectors and feed it to a highway network [27] with two layers to generate the final sentence representation vector.

## IV. SEMANTIC RETRIEVAL FRAMEWORK

*A. Framework Overview*

Fig. 6 shows the proposed semantic retrieval framework, where the encoding-based model plays a very important role and has a great impact on the overall performance. In the offline system, all questions in the FAQ set are encoded to dense real-valued vectors. Then we build an ANN vector index by using an ANN tool, such as Annoy, through which we can get the most similar vectors given a vector encoded from any new question.

TABLE I.     EXPERIMENT DATASETS

| Dataset | Language | Source | Scale (train/valid/test) | pos:neg | Overlap rate (pos/neg/avg) |
|---|---|---|---|---|---|
| Quora | English | Quora | 384,348/10,000/10,000 | 1:1.71 | 0.622/0.445/0.511 |
| LCQMC | Chinese | Baidu Knows | 238,766/8,802/12,500 | 1.35:1 | 0.771/0.530/0.668 |
| BQ | Chinese | Bank | 100,000/10,000/10,000 | 1:1 | 0.326/0.174/0.250 |
| TCS | Chinese | Telephone | 94,993/5,000/3,000 | 1:4.31 | 0.334/0.172/0.203 |

In the online system, we deploy the same module to encode the question input by the user. By inputting the vector to the ANN module, we can get top similar questions with a semantic matching score. Then the most similar question could be seen as synonymous to the user's question, so they might share the same answer. A more complicated rank module could be used following the ANN module to re-rank the top K candidates, such as interaction-based models, or ranking algorithms with hand-crafted features. However, the rank module is less important in our proposed framework than in the traditional IR-based QA frameworks.

### B. Analysis

As compared with the traditional IR-based QA frameworks (Fig. 1), our framework is less dependent on the general question analysis tools like keyword extraction. Besides, our framework removes the traditional recall module based on text search engines, which is replaced by the new recall module based on the sentence encoding and ANN technology.

## V. EXPERIMENTS

### A. Datasets

We conduct experiments on four sentence matching datasets, each comprising a large set of instances in the form of $\langle q_1, q_2, l \rangle$, where $q_1$ and $q_2$ are two questions, and $l$ is the label indicating whether they are paraphrases or not. Table I shows a brief description of these datasets. The overlap rate is a ratio of the number of common words between two sentences in a sample to the average number of them.

- Quora Question Pair dataset [15] is an open-domain English dataset derived from Quora.com. We use the same split ratio as BiMPM [9].
- LCQMC dataset [28] is an open-domain Chinese dataset collected from Baidu Knows (a popular Chinese community question answering website).
- Bank Question (BQ) dataset [29] is a specific-domain Chinese dataset for sentence semantic equivalence identification (SSEI).
- Telephone customer service (TCS) dataset is a specific-domain Chinese dataset from a real-world telephone customer service scenario, where voice speeches are converted into text using the automatic speech recognition technology.

The evaluation metric is accuracy for the Quora dataset, and F1 for other datasets.

### B. Settings of Experiments

For Quora dataset, we use the Glove-840B-300D vector as the pre-trained word embedding. The character embedding is randomly initialized with 150D and the hidden size of BiGRU is set to 300. We set $\lambda$=0.8 in the multi-task loss function. For the sentence intent classification task, we only keep the sentence clusters with question number greater than 3, and the remaining sentence clusters with question number less than or equal to 3 are regarded as a special "other" cluster. Dropout layer is also applied to the output of the attentive pooling layer, with a dropout rate of 0.1. An Adam optimizer [30] is used to optimize all the trainable weights. The learning rate is set to 4e-4 and the batch size is set to 200. When the performance of the model is no longer improved, an SGD optimizer with a learning rate of 1e-3 is used to find a better local optimum.

### C. Comparing with other methods

We compared our model with the following models:

ESIM: Enhanced Sequential Inference Model [8] is an interaction-based model for natural language inference. It uses BiLSTM to encode sentence contexts and uses the attention mechanism to calculate the information between two sentences. ESIM has shown excellent performance on the SNLI dataset.

BiMPM: Bilateral Multi-Perspective Matching model [9] is an interaction-based sentence matching model with superior performance. The model uses a BiLSTM layer to learn the sentence representation, four different types of multi-perspective matching layers to match two sentences, an additional BiLSTM layer to aggregate the matching results, and a two-layer feed-forward network for prediction.

SSE: Shortcut-Stacked Sentence Encoder [6] is an encoding-based sentence-matching model, which enhances multi-layer BiLSTM with short-cut connections. SSE has been proved to be effective in improving the performance of sentence encoder, recording state-of-the-art performance of the sentence-encoding models on Quora dataset.

DIIN: Densely Interactive Inference Network [10] is an interaction-based model for natural language inference (NLI). It hierarchically extracts semantic features from interaction space to achieve a high-level understanding of the sentence pair. It achieves state-of-the-art performance on SNLI dataset and Quora dataset.

TABLE II. EXPERIMENTAL RESULTS ON QUORA DATASET

| Models | Accuracy |
|---|---|
| Siamese-CNN | 79.60 |
| Siamese-LSTM | 82.58 |
| L.D.C | 85.55 |
| BiMPM | 88.17 |
| DIIN | **89.06** |
| ESIM | 85.0 |
| SSE | 87.8 |
| MSEM (–multi-task) | 88.11 |
| MSEM | **88.86** |

a. The first five rows are copied from [10] and the next two rows are copied from [31].

TABLE III. EXPERIMENTAL RESULTS ON LCQMC DATASET

| Models | Precision | Recall | F1 |
|---|---|---|---|
| BiLSTM-char | 67.4 | 91.0 | 77.5 |
| BiLSTM-word | 70.6 | 89.3 | 78.92 |
| BiMPM-char | 77.6 | **93.9** | 85.0 |
| BiMPM-word | 77.7 | 93.5 | 84.9 |
| ESIM | 76.54 | 93.58 | 84.21 |
| SSE | 78.23 | 93.57 | 85.21 |
| MSEM (–multi-task) | 78.23 | 93.69 | 85.27 |
| MSEM | **78.90** | 93.73 | **85.68** |

a. The first four rows are copied from [28] and the next two rows are reproduced using the SMP_toolkit [31].

TABLE IV. EXPERIMENTAL RESULTS ON BQ DATASET

| Models | Precision | Recall | F1 |
|---|---|---|---|
| Text-CNN | 67.77 | 70.64 | 69.17 |
| BiLSTM | 75.04 | 70.46 | 72.68 |
| BiMPM | 82.28 | 81.18 | 81.73 |
| DIIN | 81.58 | 81.14 | 81.36 |
| SSE | 80.16 | 80.32 | 80.24 |
| ESIM | 81.91 | 81.78 | 81.85 |
| MSEM (–multi-task) | 82.39 | 83.36 | 82.87 |
| MSEM | **82.88** | **84.36** | **83.62** |

a. The first four rows are copied from [29] and the next two rows are reproduced using the SMP_toolkit [31].

## D. Results of Experiments

The results of experiments on four sentence matching datasets are summarized as follows:

Quora dataset: Table II shows the experimental results compared with existing models on Quora dataset. Compared with SSE, the state-of-the-art encoding-base model, our MSEM model outperforms SSE by about 1% and achieves new state-of-the-art performance. In addition, our model outperforms

TABLE V. EXPERIMENTAL RESULTS ON TCS DATASET

| Models | Precision | Recall | F1 |
|---|---|---|---|
| BiMPM | 86.77 | 85.11 | 85.93 |
| ESIM | 87.19 | 87.02 | 87.11 |
| SSE | 88.54 | 87.02 | 87.78 |
| MSEM (–multi-task) | 87.38 | 88.55 | 87.96 |
| MSEM | **88.63** | **89.31** | **88.97** |

TABLE VI. ABLATION STUDY

| Models | Accuracy |
|---|---|
| MSEM | **88.86** |
| MSEM – multi-task | 88.11 |
| MSEM – ARU | 87.84 |
| MSEM – ARU + multi-head attention | 88.25 |
| MSEM – attentive pooling + max pooling | 88.35 |
| MSEM – highway network | 88.36 |
| MSEM – char embedding | 88.26 |

BiMPM and ESIM models without any sentence interaction information, and is very close to DIIN, the state-of-the-art interaction-based model, but we don't any external knowledge in our method.

LCQMC dataset: Experimental results of LCQMC dataset compared with the existing models are shown in Table III. The experimental results show that our model outperforms state-of-the-art models.

BQ dataset: BQ dataset is a specific-domain dataset with a low average overlap rate. As shown in Table IV, our model outperforms state-of-the-art models by a large margin, reaching 83.62%, recording the state-of-the-art performance.

TCS dataset: As shown in Table V, experimental results show that our MSEM model achieves the best performance. This indicates that our model is also very effective in the spoken question-answering scenario.

To sum up, experimental results show that our proposed model without multi-task learning outperforms SSE, the state-of-the-art encoding-based models, across all four datasets. And the model with multi-task learning further improved performance ranging from 0.4% to 1%. Compared with existing models, our model shows great advantages on datasets with low average overlap rate, which is known to be very common in real-world question answering scenarios.

## E. Ablation Study

The above experiments show the effectiveness of our proposed multi-task training strategy. In this section, we present the results of an ablation study on Quora dataset for evaluating the contribution of each component of the encoder, as shown in Table VI. Note that we do the significant test for each ablation experiment using the t-test ($p < 0.05$). We first study the contribution of the ARU component. The accuracy decreases

TABLE VII. EXAMPLES

| Sentences | Label | O | W | Cluster ID |
|---|---|---|---|---|
| S1: How do I build my own custom made desktop computer ? | 1 | 0 | 1 | 3840 |
| S2: How do I build a computer ? | | | | |
| S1: How many medals did India won in Olympic 2016 ? | 0 | 1 | 0 | 2303 |
| S2: How many medals will India win in 2016 Olympics ? | | | | 2736 |
| S1: Why do you believe in the afterlife ? | 0 | 1 | 0 | 0 |
| S2: How many medals will India win in 2016 Olympics ? | | | | 1717 |

a. O indicates the predicted label of MSEM without multi-task, W indicates the predicted label of MSEM. Cluster ID 0 indicates the special "other" cluster.

TABLE VIII. RESULTS OF ONLINE SYSTEM EVALUATION

| Models | Precision | Recall | F1 |
|---|---|---|---|
| Baseline | 87.88 | 64.32 | 74.28 |
| Our System | **91.56** | **79.09** | **84.87** |

from 88.86% to 87.84% after removing the ARU component. If we replace the ARU component with multi-head attention [19], the accuracy will drop to 88.25%. Next we compare the effect of attentive pooling vs max pooling. It turns out that the attentive pooling is better than max pooling. Then if we remove the highway network, the accuracy will drop to 88.36%. Finally when we remove the character-level embedding, the accuracy will drop to 88.26%. A possible reason might be that the character-level embedding can better handle the out-of-vocab (OOV) words.

*F. Online System Evaluation*

We perform an online evaluation with a telephone customer service system. We randomly select 1138 questions from the system log and send them to a baseline system and the new system, respectively. The baseline system is similar to what shown in Fig. 1, where the retrieval module is built on Elasticsearch and 30 candidate questions will be recalled and then ranked by the MSEM model. The new system is similar to what shown in Fig. 6, where an ANN module based on Annoy and a sentence-encoding module based on MSEM are adopted. We manually evaluate the returned results and measure the performance with the F1 score. As shown in Table VIII, the F1 score of the new system is 14.26% higher than the baseline system. Obviously, the semantic competence derived from the MSEM module plays a key role in the new system.

*G. Case Studies*

We perform some case studies using the Quora test set to analyze the effectiveness of the multi-task strategy. We randomly select 200 sentences with a predicted intent of non-other and manually annotate the correctness of the predicted intent. We find that the accuracy can reach 96.5%, indicating that our model can address the intent classification task pretty well. Table VII shows some examples where the MSEM model works, while the MSEM without multi-task fails. In the first example, although the text similarity between S1 and S2 is low, our model can correctly identify that they have the same intent. In the second example, S1 and S2 have high text overlap, but the model can correctly identify that they have different intents, which helps our model can better distinguish their semantics. In the third example, the model classifies S1 as "other" and S2 as "non-other", which can also help the model distinguish their semantics.

VI. CONCLUSION

In this paper, we first propose a Multi-task Sentence Encoding Model (MSEM) which addresses both the paraphrase identification task and the sentence intent classification task simultaneously, and then further propose a general semantic retrieval framework that combines the sentence encoding model and approximate nearest neighbor search technology, which can find the most similar question very quickly from all available questions in a massive QA knowledge base. We evaluated our model on several benchmark datasets. Experimental results show that our proposed method is superior to many recent sentence-matching models.